\documentclass[twoside]{article}

\usepackage[accepted]{aistats2020}

\usepackage[utf8]{inputenc} 
\usepackage[T1]{fontenc}    
\usepackage{url}            
\usepackage{amsfonts}       
\usepackage{nicefrac}       
\usepackage{microtype}      
\usepackage{amsfonts}
\usepackage{amsmath}
\usepackage{amssymb}
\usepackage{tikz}
\usetikzlibrary{fit,positioning}
\usetikzlibrary{bayesnet}
\usepackage{makecell}
\usetikzlibrary{shapes,snakes}
\usepackage{natbib}

\DeclareMathOperator*{\argmax}{argmax} 

\let\emptyset\varnothing

\usepackage{caption}
\usepackage{subcaption}

\usepackage[linguistics]{forest}
\usetikzlibrary{arrows,patterns}
\usetikzlibrary{shapes,decorations,arrows,calc,arrows.meta,fit,positioning}
\tikzset{
	bidirected/.style={Latex-Latex,dashed}
}

\DeclareMathOperator*{\argmin}{arg\,min}

\usepackage{amsthm}

\theoremstyle{definition}
\newtheorem{definition}{Definition}[section]

\usepackage[titlenumbered, ruled]{algorithm2e}

\usetikzlibrary{calc}

\usepackage{amsmath,stackengine,mathtools}
\usetikzlibrary{calc}

\newcommand{\mat}[1]{\mathbf{#1}}
\renewcommand{\vec}[1]{ \mathbf{#1} } 

\newcommand{\gp}{\mathcal{GP}}

\newcommand{\dataset}{\mathcal{D}}
\newcommand{\x}{\vec{x}}

\newcommand{\expectation}[2]{ \mathbb{E}_{#1}{\left[#2\right]} }
\newcommand{\hatexpectation}[2]{ \hat{\mathbb{E}}_{#1}{\left[#2\right]} }

\newcommand{\observations}{\mathcal{D}^{\text{O}}}
\newcommand{\true}{\mathcal{D}^{\text{I}}}

\newcommand{\DO}[2]{\text{do}\left(#1 = #2\right)}

\newcommand{\graph}{\mathcal{G}}

\newcommand{\missets}{\mathbb{M}^{\mat{C}}_{\graph, Y}}
\newcommand{\pomissets}{\mathbb{P}^{\mat{C}}_{\graph, Y}}
\newcommand{\boset}{\mathbb{B}^{\mat{C}}_{\graph, Y}}

\newcommand{\toymissets}{\mathbb{M}_{\graph, Y}}
\newcommand{\toypomissets}{\mathbb{P}_{\graph, Y}}
\newcommand{\toyboset}{\mathbb{B}_{\graph, Y}}

\newcommand{\eq}{Eq.\xspace}
\newcommand{\eqs}{Eqs.\xspace}
\newcommand{\fig}{Fig.\xspace}

\newcommand{\acro}[1]{\textsc{#1}\xspace}

\newcommand{\gptext}{\acro{gp}}

\newcommand{\rbf}{\acro{rbf}}

\newcommand{\ta}{\acro{ta}}
\newcommand{\dic}{\acro{dic}}

\newcommand{\bo}{\acro{bo}}
\newcommand{\expset}{\textbf{\acro{es}}}
\newcommand{\mab}{\acro{mab}}
\newcommand{\rl}{\acro{rl}}

\newcommand{\cbo}{\acro{cbo}}
\newcommand{\cgo}{\acro{cgo}}

\newcommand{\DAG}{\acro{dag}}

\newcommand{\pomis}{\acro{pomis}}
\newcommand{\mis}{\acro{mis}}

\newcommand{\ei}{\acro{ei}}
\newcommand{\sem}{\acro{sem}}
\newcommand{\dotext}{\textit{do}}
\newcommand{\EI}{\acro{ei}}

\newcommand{\psa}{\acro{psa}}
\newcommand{\nec}{\acro{nec}}

\begin{document}
		
\twocolumn[
	
\aistatstitle{Causal Bayesian Optimization}
	
\aistatsauthor{Virginia Aglietti \And  Xiaoyu Lu \And Andrei Paleyes \And Javier Gonz\' alez}

\aistatsaddress{
	University of Warwick\\
	The Alan Turing Institute\\ 
	\url{V.Aglietti@warwick.ac.uk}\\ \And  
	Amazon \\
	Cambridge, UK \\
	\url{luxiaoyu@amazon.com}\\ \And 
	Amazon \\
	Cambridge, UK \\ 
	 \url{paleyes@amazon.com}\\ \And 
	Amazon \\
	Cambridge, UK\\
	\url{gojav@amazon.com} }
]

\begin{abstract}
This paper studies the problem of globally optimizing a variable of interest that is part of a causal model in which a sequence of interventions can be performed. This problem arises in biology, operational research, communications and, more generally, in all fields where the goal is to optimize an output metric of a system of interconnected nodes. Our approach combines ideas from causal inference, uncertainty quantification and sequential decision making. In particular, it generalizes Bayesian optimization, which treats the input variables of the objective function as independent, to scenarios where causal information is available. We show how knowing the causal graph significantly improves the ability to reason about optimal decision making strategies decreasing the optimization cost while avoiding suboptimal solutions. We propose a new algorithm called Causal Bayesian Optimization (\cbo). \cbo automatically balances two trade-offs: the classical exploration-exploitation and the new observation-intervention, which emerges when combining real interventional data with the estimated intervention effects computed via \textit{do}-calculus. We demonstrate the practical benefits of this method in a synthetic setting and in two real-world applications. 
\end{abstract}

\vspace{-0.5cm}
\section{Introduction} \label{intro}
Decision making problems in a variety of domains, such as biological systems, modern industrial processes or social systems, require implementing interventions and manipulating variables in order to optimize an outcome of interest. For instance, in strategic planning, companies need to decide how to allocate scarce resources across different projects or business units in order to achieve performance goals. In biology, it is common to change the phenotype of organisms by acting on individual components of complex gene networks. This paper describes how to find such interventions or policies.

Focusing on a specific example, consider a setting in which $\mat{Y}$ denotes the crop yields for different agricultural products, $X$ denotes soil fumigants and $\mat{Z}=\{Z_1, Z_3, Z_4\}$ represents the eel-worm population at different times \citep{cochran1957experimental}. Given a causal graph \citep{pearl1995causal} representing the investigator's understanding of the major causal influences among the variables (\fig \ref{fig:dag_examples}, left), she aims at finding the highest yielding intervention in a limited number of seasons and subject to a budget constraint. Each intervention has a cost which is determined by the manipulated variables' costs and the implemented intervention levels.

In order to solve this problem, the investigator could resort to Bayesian Optimization (\bo). \bo is an efficient heuristic to optimize objective functions whose evaluations are costly and when no explicit functional form is available \citep{jones1998efficient}. In the example above,  \bo would try to find the global optima by making a series of function evaluations in which all variables  are manipulated. \bo would thus break the dependency structure existing among $X$ and $\mat{Z}$, potentially leading to suboptimal solutions. Indeed, as described later in detail, depending on the structural relationships between variables, intervening on a subgroup might lead to a propagation of effects in the causal graph and a higher final yield. In addition, intervening on all variables is cost-ineffective in cases when the same yield can be obtained by setting only a subgroup of them.

\subsection{Approach and Contributions}
The framework proposed in this work combines \bo and causal inference, offering a novel approach for decision making under uncertainty.
Probabilistic causal models are commonly used in disciplines where explicit experimentation may be difficult such as social science or economics. In particular, the \dotext-calculus \citep{pearl1995causal} relates observational distributions to interventional ones. It allows to predict the effect of an intervention without explicitly performing it and by solely using observational data.
We develop a model which integrates observational and interventional data so as to further reduce the uncertainty around the optimal value and the number of interventions required to find it. Particularly, we make the following contributions:
\vspace{-0.2cm}
\begin{itemize}
	\item We formulate a new class of optimization problems called \emph{Causal Global Optimization} (\cgo) where the causal structure existing among the input variables is accounted for in the objective functions.
	 
	\item We solve \cgo problems by combining ideas from \bo and causal calculus. We propose a Gaussian process (\gptext) surrogate model, the causal \gptext, that integrates observational and interventional data via the definition of a causal prior distribution computed through \dotext-calculus. 
	
	\item We propose an acquisition function, the causal expected improvement (\ei), which drives the exploration of different interventions. 
		
	\item We develop an algorithm, henceforth named \emph{Causal Bayesian Optimization} (\cbo), that exploits the structural characteristics of the graph, the causal \ei and the proposed \gptext prior to find an optimal intervention. In doing that, it balances the emerging trade-off between observation and intervention via an $\epsilon$-greedy policy.   
	
	\item We show the benefits of the proposed approach in a variety of experimental settings featuring different dependency structures, unobserved confounders and non-manipulative variables.
\end{itemize}

\subsection{Related Work}
 While there exists an extensive literature on \bo (see \citet{shahriari2015taking} for a review) and causality \citep{guo2018survey}, the literature on causal decision making is limited. 
Recent works have focused on multi-armed bandit (\mab) problems and reinforcement learning (\rl) settings  where actions or arms correspond to interventions on an arbitrary causal graph and there exists complex links between the agent’s decisions and the received rewards. \cite{bareinboim2015bandits} and  \cite{lu2018deconfounding} focus on settings with unobserved confounders. \cite{lee2018structural} identify a set of possibly-optimal arms that an agent should play in order to maximize its expected reward in a \mab problem. \cite{lee2019structural} extend this work to graphs with non-manipulable variables. \cite{lattimore2016causal} study a specific family of \mab problems called parallel bandit problems.  
Finally, \cite{ortega2014generalized} focus on causal discovery in Causal \mab. In the \rl literature, \cite{buesing2018woulda} leverage structural causal models for counterfactual evaluation of arbitrary policies on individual off-policy episodes. \cite{foerster2018counterfactual} focus on the multi-agents setting and propose a framework in which each agent learns from a shaped reward that compares the global reward to the counterfactual reward received when that agent’s action is replaced with a default action.

\section{Background and Problem Statement}
\begin{figure}
	\centering
		\vspace{-0.4cm}
	\begin{subfigure}[b]{0.4\textwidth}
	\hspace{-0.4cm}
	\centering
		\begin{tikzpicture}[shorten >=1pt, node distance=1cm, baseline=4ex]
		\node[obs](Z3){$Z_3$}; 
		\node[obs, below= of Z3, yshift=0.5cm](Z2){$Z_2$}; 
		\node[obs, below= of Z2,yshift=0.5cm](X){$X$}; 
		\node[obs, right= of Z2, xshift=-0.3cm](Y){$Y$}; 
		\node[obs, left= of Z2, xshift=0.3cm](Z1){$Z_1$}; 
		\edge {Z2}{Z3} ; %
		\edge {X} {Z2} ; %
		\edge {Z3} {Y} ; %
		\edge {X} {Y} ; %
		\edge {Z2} {Y} ; %
		\edge {Z1} {Z2} ; %
		\path[bidirected] (Z3) edge[bend right=10] (Z1);
		\path[bidirected] (Z1) edge[bend right=10] (X);
		\end{tikzpicture}
	\begin{subfigure}[b]{0.3\textwidth}
		\hspace{0.5cm}
		\begin{tikzpicture}[shorten >=1pt, node distance=2mm, on grid, baseline=0ex]
		\node[obs] (X1) {$X_1$} ; %
		\node[obs, below = of X1, yshift=-0.9cm] (X100) {$X_{100}$} ; %

		\node[obs, right = of X1] (Z1) {$Z_1$} ; %
		\node[obs, below = of Z1, yshift=-0.9cm] (Z100) {$Z_{100}$} ; %
		
		\node[obs, below right = of X100, xshift=-0.2cm, yshift=-0.5cm] (Y) {$Y$} ; %
		
		\node[yshift=-0.1cm] at ($(X1)!.4!(X100)$) (node1) {$\mat{\dots}$};
		\node[yshift=-0.1cm] at ($(Z1)!.4!(Z100)$) (node2) {$\mat{\dots}$};
			
		\edge {X1} {node1} ; %
		\edge {Z1} {node2} ; %
		\edge {node1} {X100} ; %
		\edge {node2} {Z100} ; %
		\edge {X100} {Y} ; %
		\edge {Z100} {Y} ; %
		\end{tikzpicture}
	\end{subfigure}%

\end{subfigure}
	\caption{Examples of causal graphs. Nodes denote variables, arrows represent causal effects and dashed edges indicate unobserved confounders. \emph{Left}: Yield optimization example. $Y$ is the crop yield, $X$ denotes soil fumigants and $\mat{Z}$ represents the eel-worm population. \emph{Right}: A $200$-dimensional optimization problem with causal intrinsic dimensionality equal to $2$. }
	\vspace{-0.3cm}
	\label{fig:dag_examples}
\end{figure}
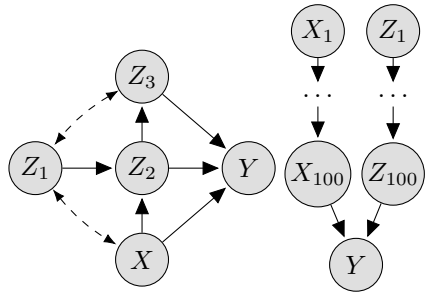

In this paper, random variables and observations are denoted in upper case  and lower case respectively. Vectors are represented in bold. $do(\mat{X} = \mat{x})$ represents an intervention on $\mat{X}$ whose value is set to $\mat{x}$. $P(\mat{Y}|\mat{X} = \mat{x})$ represents an \emph{observational distribution} and $P(\mat{Y}|\DO{\mat{X}}{\mat{x}})$ represents an \emph{interventional distribution}. $\mathcal{D}^O$ and $\mathcal{D}^I$ denote observational and interventional datasets respectively. 

\subsection{Structural Equation Models (\sem) and \dotext-calculus}
Consider a probabilistic causal model  \citep{pearl2000causality} consisting of a directed acyclic graph $\graph$ (\DAG) and a four-tuple  $\langle \mat{U}, \mat{V}, F, P(\mat{U})\rangle$, where $\mat{U}$ is a set of independent \emph{exogenous} background variables distributed according to the probability distribution $P(\mat{U})$, $\mat{V}$ is a set of observed \emph{endogenous} variables and $F = \{f_1, \dots, f_{|\mat{V}|}\}$ is a set of functions such that $v_i = f_i(pa_i, u_i)$ with $pa_i$ denoting the parents of $V_i$. $\graph$ encodes our knowledge of the existing causal mechanisms among $\mat{V}$. Within $\mat{V}$, we distinguish between three different types of variables: non-manipulative variables $\mat{C}$, which cannot be modified, treatment variables $\mat{X}$ that can be set to specific values and output variables $\mat{Y}$ that represents the agent's outcomes of interest. Given the conditional independence relationships encoded in $\graph$, we can exploit the causal Markov condition \citep{pearl2000causality} to write the joint observational distribution as $P(\mat{V}) = \prod_{i=1}^{|\mat{V}|} p(V_i|pa_i)$. The interventional distribution for two disjoint sets in $\mat{V}$, say $\mat{X}$ and $\mat{Y}$, is $P(\mat{Y}| \DO{\mat{X}}{\x})$. This is the distribution of $\mat{Y}$ obtained by intervening on $\mat{X}$ and fixing its value to $\x$ in the data generating mechanism, irrespective of the values of its parents and keeping $\mat{C}$ unchanged. Note the difference between $P(\mat{Y}| \DO{\mat{X}}{\x})$ that requires ``real'' interventions and $P(\mat{Y}| \mat{X = \x})$ that only requires ``observing'' the system. In this work we assume $\graph$ to be known. Causal discovery \citep{glymour2019review} is a complex topic and analysing what happens when the graph is unknown goes beyond the scope of this paper. We leave this for future work.

\textit{Do}-calculus offers a powerful tool to estimate interventional distributions and causal effects from observational distributions. If the causal effects are identifiable, we can apply the three rules of \dotext-calculus to link interventional distributions with observational distributions which can be approximated with e.g. Monte Carlo estimates. The first formula, called "back-door adjustment", applies to cases in which we have a set of observed confounders between cause and effect. For instance, in \fig \ref{fig:dag_examples} (left),  $p(y|\DO{Z_2}{z_2})$ can be computed by adjusting for $Z_1$ and $X$:
\begin{align*}
	p(y|\DO{Z_2}{z_2}) &= \int_{x,z_1} \hspace{-0.4cm} p(y|Z_2=z_2, z_1, x)p(x, z_1)dx d z_1 \text{.}
\end{align*}
The second formula, called "front-door adjustment" applies to cases with unobserved confounders. When computing $p(y|\DO{X}{x})$ in \fig \ref{fig:dag_examples} (left), the adjustment is given by:
\begin{align*}
p(y|\DO{X}{x}) &= \int_{z_1, z_2, z_3} \hspace{-0.9cm} p(y|z_2,z_3,X=x) p(z_2|z_1,X=x) \\ & \times \int_{x'}\hspace{-0.13cm}p(z_3|z_1,z_2,x') p(z_1,x') dx' dz_1dz_2dz_3\text{.}
\end{align*}
When the above integrals are not tractable, observational data can be used to get a Monte Carlo estimate $\hat{P}(\mat{Y}|\DO{\mat{X}}{\x})\approx P(\mat{Y}|\DO{\mat{X}}{\x})$, which is consistent when the number of samples drawn from $P(\mat{V})$ is sufficiently large.

\begin{figure}
	\centering
	\vspace{-0.4cm}
	\begin{subfigure}[b]{0.3\textwidth}
	\centering
		\begin{tikzpicture}[shorten >=1pt, node distance=1cm]
		\node[obs](B){$B$}; %
		\node[obs, right = of B, pattern=dots] (C) {$C$} ; 
		\node[obs, above = of C, yshift=-0.4cm, pattern=dots] (A) {$A$} ; %
		\node[obs ,right= of A] (E) {$E$} ; %
		\node[obs, right = of C] (D) {$D$} ; %
		\node[obs, above right = of D, yshift=-0.3cm] (Y) {$Y$} ; %
		\edge {C} {E} ; %
		\edge {A} {E} ;
		\edge {B} {C} ;
		\edge {C} {D} ;
		\edge {D} {Y} ;
		\edge {E} {Y} ;
		\path[bidirected] (B) edge[bend left=10] (Y);
		\path[bidirected] (A) edge[bend left=55] (Y);
		\end{tikzpicture}
		
\vspace{0.1cm}
\hspace{-4.5cm}
	\begin{subfigure}[b]{0.3\textwidth}
	\centering
	\begin{tikzpicture}[shorten >=1pt, node distance=1mm, on grid]
	\node[obs] (X_1) {$B$} ; %
	\node[obs, right = of X_1, xshift=0.2cm] (X_2) {$E$} ; %
	\node[obs, right = of X_2, xshift=0.2cm] (X_3) {$D$} ; %
	%
	\node[obs, right = of X_3, pattern=dots] (C) {$A$} ; %
	\node[obs, right = of C, xshift=0.2cm, pattern=dots] (C_2) {$C$} ; %
	\node[obs, below right=of X_3, xshift=-0.7cm, yshift=-0.3cm](Y){$Y$}; %
	\edge {X_1} {Y} ; %
	\edge {X_2} {Y} ; %
	\edge {C} {Y} ; %
	\edge {C_2} {Y} ; %
	\edge {X_3} {Y} ; %
	\end{tikzpicture}
\end{subfigure}%
	
\end{subfigure}
	\caption{\DAG representation of a \cgo problem (top) and the \DAG considered when using \bo (bottom) to address the same problem. Shaded nodes represent $\mat{X}$ while dotted nodes give $\mat{C}$. Dashed edges indicate unobserved confounders.}
	\label{fig:dag_related_methods}
	\vspace{-0.4cm}
\end{figure}
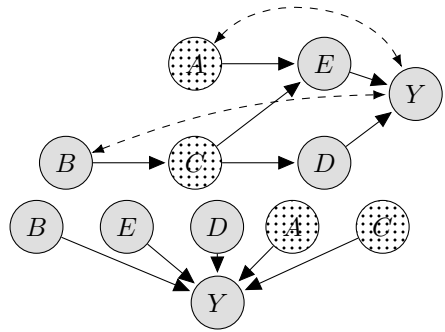

\subsection{Problem Setup}
A novel class of global optimization problems called \emph{Causal Global Optimization} (\cgo) is introduced in this section. Given $\graph$ and $\langle \mat{U}, \mat{V}, F, P(\mat{U})\rangle$, the goal is to select the set of intervention variables $\mat{X}^\star_s$ and intervention levels $\x^\star_s$ that optimize the expected target outcomes $\mat{Y}$. Formally, the goal is to find:
\begin{align}
\mat{X}^\star_s, \x^\star_s = \argmin_{\mat{X}_s \in \mathcal{P}(\mat{X}),  \x_s \in D(\mat{X}_s)}	\expectation{P(\mat{Y}| \DO{\mat{X}_s}{\x_s})}{\mat{Y}} \text{,}
\label{eq:goal_maximisation}
\end{align}
where $\mathcal{P}(\mat{X})$ is the power set of  $\mat{X}$ and $D(\mat{X}_s) = \times_{X \in \mat{X}_s}(D(\mat{X}))$ with $D(\mat{X})$ 
denoting the interventional domain of $\mat{X}$ and the expectation is computed according to the interventional distribution. For notational convenience, we denote $\expectation{P(\mat{Y}| \DO{\mat{X}_s}{\x_s})}{\mat{Y}} \doteq \expectation{}{\mat{Y}| \DO{\mat{X}_s}{\x_s}}$ and $\expectation{\hat{P}(\mat{Y}| \DO{\mat{X}_s}{\x_s})}{\mat{Y}} \doteq \hatexpectation{}{\mat{Y}| \DO{\mat{X}_s}{\x_s}}$. The optimal subset of intervention variables $\mat{X}_s$ belongs to $\mathcal{P}(\mat{X})$ which includes the empty set $\emptyset$ and $\mat{X}$ itself. When $\mat{X}_s = \emptyset$, no intervention is implemented in the system and the target expected values correspond to the observational expected outcomes.  When $\mat{X}_s = \mat{X}$, all variables are intervened upon except for the context variables $\mat{C}$ that can only be observed.
 
The problem given in \eq \eqref{eq:goal_maximisation} is challenging because of two reasons. 
Firstly, the cardinality of $\mathcal{P}(\mat{X})$ grows exponentially with $|\mat{X}|$ and finding the optimal set requires, in principle, a combinatorial search.  
Secondly, for every set $\mat{X}_s$, finding $\x^\star_s$ requires evaluating the objective function and thus implementing interventions in the system several times. In most settings, the number of function evaluations, whose cost is assumed to be given by some cost function $Co(\mat{X}_s, \x_s)$, needs to be kept low. We thus want to find the optimal configuration with the minimal cost, $\sum_{i=1}^T Co(\mat{X}_s, \x_i)$ for a sequence of interventions $1, \dots, T$.

\subsection{Connections and Generalisations}
\label{connection}
\paragraph{Bayesian Optimization:} 
Consider the \DAG in \fig \ref{fig:dag_related_methods} (top). The problem in \eq \eqref{eq:goal_maximisation} can be solved through the \bo method which breaks the input variables dependencies (\fig \ref{fig:dag_related_methods} bottom) and intervenes simultaneously on all of them thus setting $\mat{X}_s = \mat{X}$. To this aim, \bo considers a surrogate probabilistic model for the objective function $\mat{Y} = f(\mat{X}$) which is given by a \gptext $p(f) = \gp (m, k)$ with mean function $m$ and covariance function $k$. Given a dataset $\mathcal{D}_n = \{\x_i, y_i\}_{i=1}^n$, the posterior distribution of $f$ under Gaussian likelihood is also a \gptext with
posterior mean and variance given by $m_n(\x) = \vec{k}_n(\x)^T[\mat{K}_n + \sigma^2_n\mat{I}]^{-1}\mat{y}_n$ and $\sigma^2_n(\x) = k(\x, \x) - \mat{k}_n(\x)^T[\mat{K}_n + \sigma^2_n]^{-1}\mat{k}_n(\x)$ where $\mat{K}_n$ is the matrix such that $(\mat{K}_n)_{ij} = k(\x_i, \x_j)$, $\mat{k}_n(\x) = [k(x_1, \x), ..., k(x_n, \x)]^T$ \citep{rasmussen2003gaussian}  and $\x$ is the point where the \gptext is evaluated. This posterior is used to form the acquisition function $\alpha(\x, \mathcal{D}_n)$, which can be chosen among popular acquisition functions such as expected improvement or entropy search. The next evaluation is placed at the (numerically estimated) global maximum of $\alpha(\x, \mathcal{D}_n)$. 

\paragraph{Causal Dimensionality:}
It is well known \citep{wang2016bayesian} that the performance of standard \bo algorithms deteriorates in high dimensional problems  as the number of evaluations needed to find the global optimum increases exponentially with the space dimensionality. Interestingly, knowing the causal graph allows to reason about the effective dimensionality of the problem. We formalize this idea by defining the notion of \emph{causal intrinsic dimensionality}:
\theoremstyle{definition}
\begin{definition}{}
	The causal intrinsic dimensionality of a causal function $\expectation{P(Y| \DO{\mat{X}}{\mat{x}})}{Y}$ is given by the number of parents of the target variable, that is $|Pa(Y)|$. 
\end{definition}
In \fig \ref{fig:dag_examples} (right), the input space dimensionality is 200. However, $\expectation{}{Y|\text{do}(X_1,\dots,X_{100}, Z_1, \dots, Z_{100})} = \expectation{}{Y|\text{do}(X_{100},Z_{100})}$, thus $X_{100}$ and $Z_{100}$ are the only 2 relevant variables and the intrinsic dimensionality of the problem is 2. For the general problem in \eq \eqref{eq:goal_maximisation} we have $\expectation{}{\mat{Y}|  \text{do} (\mat{X}) } = \expectation{}{\mat{Y}| \text{do} (Pa(\mat{Y}))}$.

Related to the concept of causal dimensionality, \cite{wang2016bayesian} proposed to perform Bayesian optimization in a low-dimensional space which reflects the \emph{intrinsic dimensionality} of a function. Provided that the objective function has low intrinsic dimensionality,  \cite{wang2016bayesian} use random embeddings to reduce the problem dimensionality without knowing which dimensions are important. 
This idea can be formalized and made explicit by taking a causal perspective on the optimization problem. The causal graph allows to determine not only if the function has low intrinsic dimensionality but also to identify which dimensions are important.

\paragraph{Causal Bandits:}
There is a significant link between our problem setup and the settings tackled by causal \mab algorithms. Causal \mab algorithms interpret decisions as interventions, target a causal effect function and account for complex dependency structure between actions which are encoded in the causal graph. Indeed, when all intervention variables $\mat{X}$ are binary, the \cgo setting reduces to the causal \mab setting. 
However, \eq \eqref{eq:goal_maximisation} gives a more general formulation of the problem where variables can be continuous or categorical and, more importantly, where the intervention values need to be determined together with the intervention set. 
\begin{figure*}
	\begin{minipage}{.25\textwidth}
		\vspace{-0.3cm}
	 \begin{subfigure}[t]{0.2\textwidth}
		\tikz{ %
			\node[obs] (X) {$X$} ; %
			\node[obs, right = of X, xshift=0.2cm] (Z) {$Z$} ; %
			\node[obs, right = of Z] (Y) {$Y$} ; %
			\edge {X} {Z} ; %
			\edge {Z} {Y} ; %
				\node[rectangle, below=-0.2cm, text width=3cm, below = of Z, yshift=1.2cm, xshift=-0.9cm] (eq) {
				\begin{align*}
				X &=  \epsilon_X \\
				Z &= \exp(-X) + \epsilon_Z\\
				Y & = \text{cos}(Z) - \exp(-\frac{Z}{20}) + \epsilon_Y
				\end{align*}} ; %
			\node[rectangle, draw, below=-0.2cm, text width=4cm, below = of eq, yshift=1.cm, xshift=0.9cm] (eq2) {$\toymissets =  \{\emptyset, \{X\}, \{Z\}\}$ \\ $\toypomissets = \{\{Z\}\}$ \\ $\toyboset = \{\{X,Z\}\}$} ; %
		}
	\end{subfigure}
	\end{minipage}
\begin{minipage}{0.4\textwidth}
	\centering
	\includegraphics[width=0.9\linewidth]{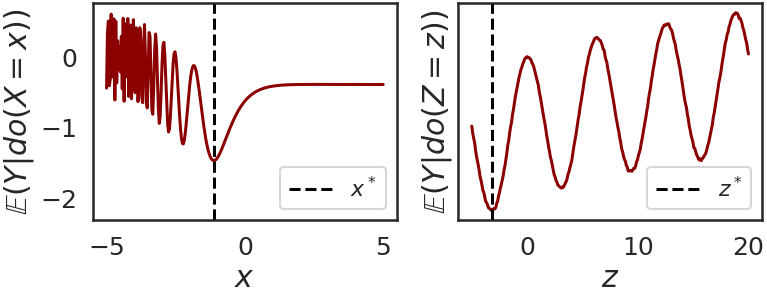}
	\includegraphics[width=0.6\linewidth]{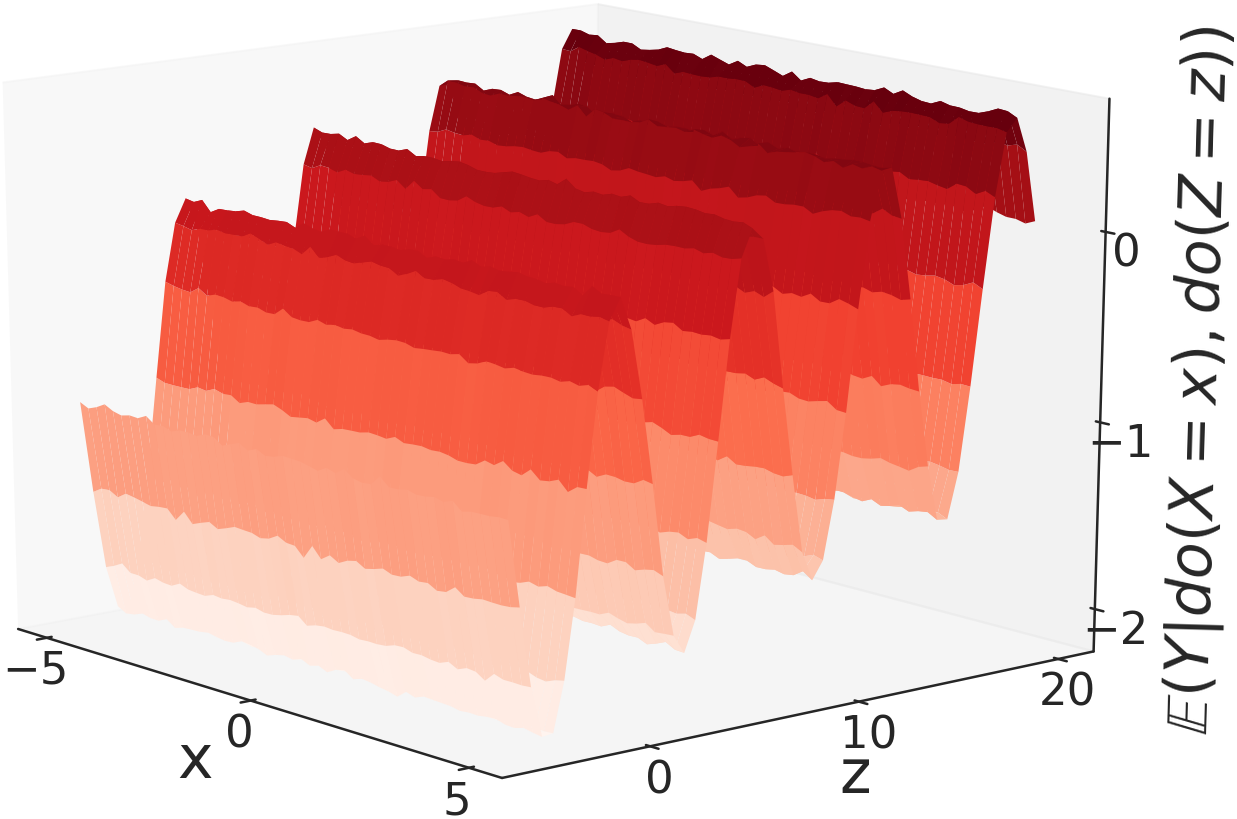} 
\end{minipage}%
\begin{minipage}{0.35\textwidth}
	\vspace{-0.3cm}
\centering
\includegraphics[width=1.0\linewidth]{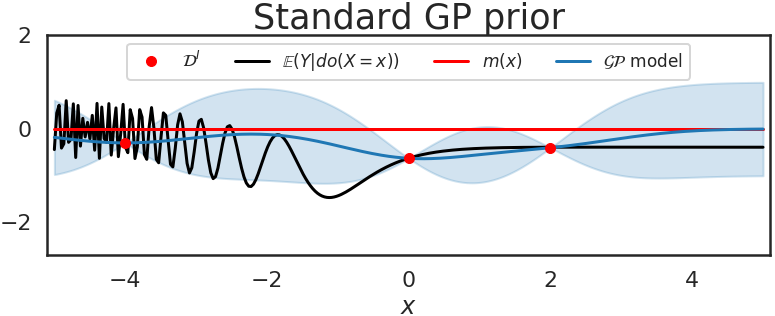} \\
\includegraphics[width=1.\linewidth]{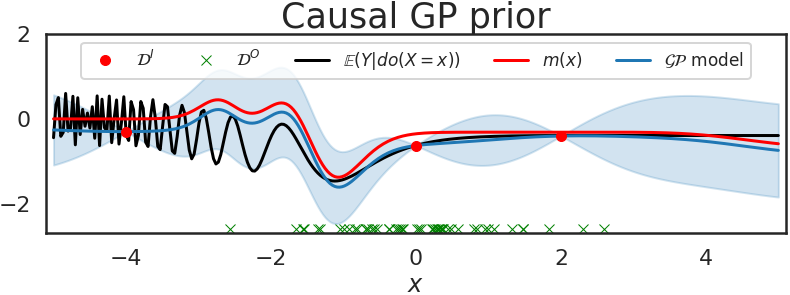} 
\end{minipage}
    \caption{Toy example illustrating the elements of \cbo. \emph{Left panel}: \DAG, \sem and optimal sets. \emph{Central panel}: Objective functions for different intervention sets. \emph{Right panel}: Posterior \gptext obtained with two different prior.}
\label{fig:toy_1}
\vspace{-0.3cm}
\end{figure*}

\section{Methodology}
This section details a new methodology, which we call \emph{Causal Bayesian Optimization}, addressing the problem in \eq \eqref{eq:goal_maximisation}. The elements of this approach are: (i) an exploration set (\S \ref{optimal_set}) determining a set of variables which is worth intervening on based on the topology of $\graph$, (ii) a surrogate model (\S \ref{causal_gp}), called Causal \gptext model, that enables the integration of observational and interventional data, (iii) an acquisition function (\S \ref{causal_acquisition}) solving the exploration/exploitation trade off \emph{across} interventions, (iv) an $\epsilon$-greedy policy (\S \ref{greedy_policy}) solving the observation/intervention trade-off.
This paper considers settings where a data set $\observations = \{(\vec{v}^n,y^n)\}_{n=1}^N$ from an observational study is available. Here $\vec{v}^n \in \mathbb{R}^{|\vec{V}|}$, $y^n \in \mathbb{R}$ and the joint distribution follows the conditional independence assumptions encoded in $\graph$. 
In the following discussion we consider a single output and leave the integration of our framework with multi-output schemes as future work.

\subsection{Selecting the Optimal Exploration Set}
\label{optimal_set}
A naive approach to find $\mat{X}_s^\star$ would be to explore the $2^{|\mat{X}|}$ sets in $\mathcal{P}(\mat{X})$. Albeit this is a valid strategy, its complexity grows exponentially with $|\mat{X}|$. However, exploiting the rules of \dotext-calculus and the partial orders among subsets, \citet{lee2018structural} identify invariances in the interventional space and potentially optimal intervention set which we define below.

 \begin{definition}{\textbf{Minimal Intervention set (\mis).}}
Given $\langle \graph, \mat{Y}, \mat{X},\mat{C} \rangle$, a set of variables $\mat{X}_s \in \mathcal{P}(\mat{X})$ is said to be a \mis if there is no $\mat{X}'_s \subset \mat{X}_s$ such that $\expectation{}{Y| \DO{\mat{X}_s }{\x_s}} = \expectation{}{Y|\DO{\mat{X}'_s }{\x'_s}}$. 
\end{definition}

 We denote by $\mathbb{M}^{\mat{C}}_{\graph, Y}$ the set of \mis{s} for $\langle \graph, \mat{Y}, \mat{X},\mat{C} \rangle$ where each \mis represents a set of variables that is worth intervening on. When $\mat{C} = \emptyset$, we use  $\mathbb{M}_{\graph, Y}$. Incorporating into \mis the partial orderedness among subsets of $\mathcal{P}(\mat{X})$ we define the so-called \pomis.

 \begin{definition}{\textbf{Possibly-Optimal Minimal Intervention set (\pomis)}.}
	Given $\langle \graph, \mat{Y}, \mat{X},\mat{C} \rangle$, let $\mat{X}_s \in \mathbb{M}^{\mat{C}}_{\graph, Y}$. $\mat{X}_s$ is a \pomis if there exists a \sem conforming to $\graph$ such that $\expectation{}{Y| \DO{\mat{X}_s }{\x^*}} > \forall_{\mat{W} \in \mathbb{M}^{\mat{C}}_{\graph, Y} \setminus \mat{X}_s} \expectation{}{Y| \DO{\mat{W} }{\mat{w}^*}}$ where $\x^*$ and $\mat{w}^*$ denote the optimal intervention values.
\end{definition}
We denote by $\mathbb{P}^{\mat{C}}_{\graph, Y}$ the set of \pomis for $\langle \graph, \mat{Y}, \mat{X}, \mat{C} \rangle$ where each \pomis represents a variable on which intervening always improves $Y$ with respect to the remaining elements in $\missets$. For completeness, we also use $\boset$ to denote the unique set on which \bo performs interventions that includes all manipulative variables $\mat{X}$. In \fig \ref{fig:toy_1} we give an example in which $|\mathbb{M}_{\graph, Y}| < |\mathcal{P}(\mat{X})|$ and intervening on $\mathbb{B}_{\graph, Y}$ is suboptimal. Indeed, the central panel shows how $\expectation{}{Y|\DO{X}{x},\DO{Z}{z}} = \expectation{}{Y|\DO{Z}{z}}$ and the causal intrinsic dimensionality of this problem is $|Pa(Y)|= 1$. In addition, $\expectation{}{Y| \DO{X}{x^*}} > \expectation{}{Y| \DO{Z}{z^*}}$ thus $Z$ is optimal with respect to $X$ and is the only set in $\mathbb{P}_{\graph, Y}$. 
For notational convenience, we refer to the exploration set, which can be $\mathbb{M}^{\mat{C}}_{\graph, Y}$  or $\pomissets$, as $\expset$.
The choice of the $\expset$ depends on the causal graph and the next sections are agnostic to the choice of the $\expset$.

\subsection{Causal \gptext Model}
\label{causal_gp}
To integrate experimental and observational data, for each $\mat{X}_s \in \expset $, we place a \gptext prior on $f(\mat{x}_s) = \expectation{}{Y| \DO{\mat{X}_s}{\x_s}}$ with prior mean and kernel function computed via do-calculus:
\begin{align}
f(\x_s) &\sim \mathcal{GP}(m(\x_s), k_C(\x_s, \x'_s))  \label{eq:gp_model}\\
m(\x_s) &=\hatexpectation{}{Y|\DO{\mat{X}_s}{\x_s}} \label{eq:mean_function} \\
k_C(\x_s, \x'_s) &= k_{RBF}(\x_s, \x'_s) + \sigma(\x_s)\sigma(\x'_s)
\label{eq:model_formulation}
\end{align}
where $\sigma(\x_s) = \sqrt{\hat{\mathbb{V}} (Y|\DO{\mat{X}_s}{\x_s})}$ with $\hat{\mathbb{V}} $ representing the variance estimated from observational data and $k_{RBF}$ representing the radial basis function kernel defined as $k_{RBF}(\x_s, \x'_s) := \exp(-\frac{||\x_s - \x'_s ||^2}{2l^2})$.
 \fig \ref{fig:toy_1} (right) illustrates the difference between the posterior \gptext distribution obtained with a zero mean prior distribution and \rbf kernel (upper) and with the proposed causal prior (lower). 
Notice how the prior mean function captures the behaviour on the target function in area where observations are available (crosses at the bottom) despite the lack of interventional data. In addition, the posterior variance is higher in area where  $\sqrt{\hat{\mathbb{V}} (Y|\DO{\mat{X}_s}{\x_s})}$ is inflated due to the lack of observational data, which enables proper calculation of the uncertainties about the causal effects.

\begin{figure*}[htbp]
    \begin{subfigure}[b]{0.33\textwidth}
\includegraphics[width=\textwidth]{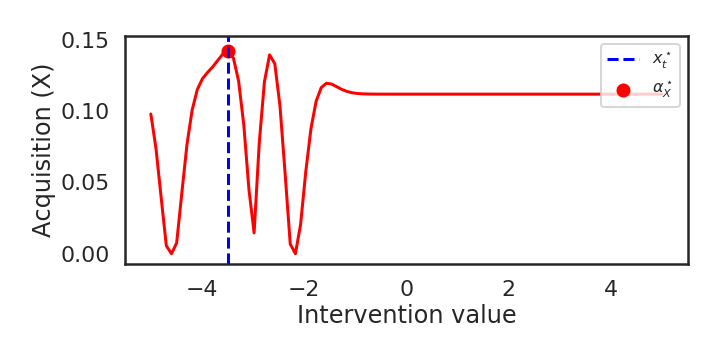} 
    \end{subfigure}
     \begin{subfigure}[b]{0.33\textwidth}
\includegraphics[width=\textwidth]{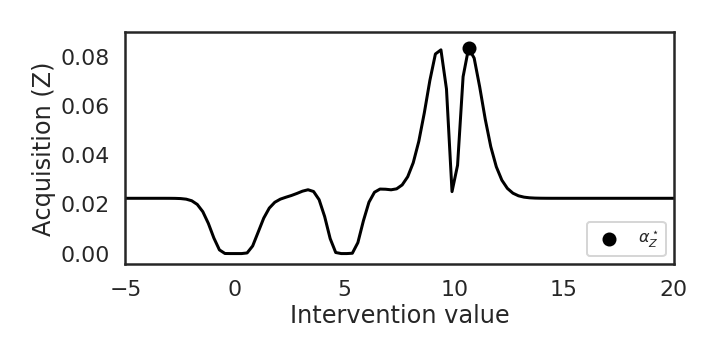} 
    \end{subfigure}
	\begin{subfigure}[b]{0.33\textwidth}
		\includegraphics[width=\textwidth]{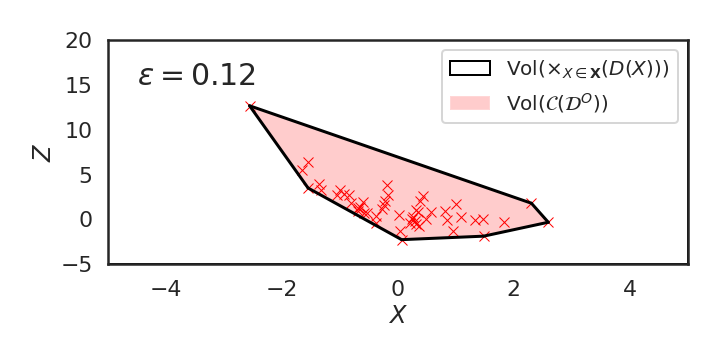} 
	\end{subfigure}
    \caption{Toy example. \emph{Left and central panels}: Acquisition functions for the variables in $\missets$. The dashed blue line gives the next optimal intervention. \emph{Right panel}: Convex hull in the $X$-$Z$ space used to calculate $\epsilon$. Red crosses give $\dataset^O$ while the boundaries of the plot correspond to the interventional domain.}
\label{fig:toy_2}
\vspace{-0.3cm}
\end{figure*}

\subsection{Causal Acquisition Function} 
\label{causal_acquisition}
For each $\mat{X}_s \in \expset $, we define the acquisition function as the expected improvement (\EI) with respect to the current best observed interventional setting across all sets in \expset. 
At every step of the optimization, denote by $y_s = \expectation{}{Y| \DO{\mat{X}_s}{\x}}$  and $y^\star$ the optimal value of $y_s\text{, } s=1, \dots, |\expset|$ observed so far. The \EI is given by:
\begin{align}
	EI^s(\x) = \expectation{p(y_s)}{\text{max}(y_s-y^\star, 0)}/Co(\mat{x}).
	\label{eq:causal_acquisition}
\end{align}
	
Let $\alpha_1, \dots, \alpha_{|\expset|}$ be solutions of the optimization of $EI^s(\x)$ for each set in $\expset$ and $\alpha^\star := \max \{\alpha_1, \dots, \alpha_{|\expset|}\}$. The best new intervention set is given by $s^\star = \argmax_{s \in \{1, \cdots, |\expset|\}} \alpha_s.$ Therefore, the set-value pair to intervene on is $(s^\star, \alpha^\star)$. \fig \ref{fig:toy_2} (left and center) shows the acquisition functions for $\expset=\missets$ in the toy example. The new intervention is selected by comparing the maxima of the acquisition functions across interventions (red and back dots). 

\subsection{$\epsilon$-greedy Policy}
\label{greedy_policy}
For some graph structures, such as in Fig \ref{fig:dag_related_methods} (top), the empty set, which represents the observational case, is part of \expset. A mechanism in the optimization process is then needed to observe the system when that is the optimal strategy. Inspired by the $\epsilon$-greedy policies in \rl \citep{tokic2010adaptive}, we propose an $\epsilon$-greedy approach where $\epsilon$ determines the probability of observing and trades off exploration and exploitation. The value of $\epsilon$, which is a parameter of \cbo, can be selected in several different ways and needs to balance the \emph{observation-intervention trade-off} emerging in \cbo. On the one hand, collecting observational data allows to reliably estimate causal effects via \dotext-calculus. On the other hand, computing consistent causal effects for values outside of the observational range, requires intervening. The agent needs to find the right combination of the two actions so as to exploit observational information while intervening in areas where the uncertainty is higher. In this paper, we define $\epsilon$ as:
\begin{align}
\epsilon =
\frac{\mathsf{Vol}(\mathcal{C}(\mathcal{D}^O))}{\mathsf{Vol}(\times_{X \in \mat{X}}(D(X)))} \times \frac{N}{N_{\text{max}}},
\end{align}
where $\mathsf{Vol}(\mathcal{C}(\mathcal{D}^O))$ represents the volume of the convex hull for the observational data and $\mathsf{Vol}(\times_{X \in \mat{X}}(D(X)))$ gives the volume of the interventional domain (see \fig \ref{fig:toy_2} (right)). $N_{\text{max}}$ represents the maximum number of observations the agent is willing to collect and $N$ is the current size of $\dataset^O$.
When $\mathsf{Vol}(\mathcal{C}(\mathcal{D}^O))$ is small with respect to $N$, the interventional space is bigger than the observational space. We thus intervene and explore part of the interventional space not covered by $\observations$. On the contrary, if $\mathsf{Vol}(\mathcal{C}(\mathcal{D}^O))$ is large with respect to $N$, we 	obtain consistent estimates of the causal effects by collecting more observations. We thus observe and update the prior \gptext in  \eqs \eqref{eq:mean_function} - \eqref{eq:model_formulation}.

Other $\epsilon$-greedy policies can be formulated in order to solve the trade-off differently.
For instance, the agent could define an adaptive $\epsilon$ which favours observations in the first stages of the optimization procedure and interventions as $N$ increases. Alternatively, the value of $\epsilon$ could depend on the agent's budget and favours interventions when their cost is low. 

\subsection{The \cbo Algorithm}
We give the complete \cbo algorithm in Alg. \ref{alg:ceo}). 
The time complexity of \cbo is dominated by algebraic operations on $k_C(\x_s, \x'_s)$ which are $\mathcal{O}(P^3)$ where $P$ denotes the number of function evaluations of the \bo algorithm. The space complexity is dominated by storing $k_C(\x_s, \x'_s)$ which is $\mathcal{O}(P^2)$. Given the acquisition function and the surrogate model, the theoretical guarantees of \cbo are limited and follow directly from the theoretical properties of \textit{do}-calculus. However, one could extend \cbo to use a \gptext-\acro{ucb} acquisition function for which a cumulative regret bound has been derived \citep{srinivas2012information}. We leave this for future work. 

\begin{algorithm}[t]
	\SetKwInOut{Input}{input}
	\SetKwInOut{Output}{output}
	\KwData{$\observations$, $\true$, $\graph$, \expset, number of steps $T$ }
	\KwResult{$\mat{X}^\star_s, \x^\star_s, \hatexpectation{}{\mat{Y}^\star|\DO{\mat{X}^*_s}{\x^\star_s}}$}
	
	\textbf{Initialise}: Set $\true_0 = \true$ and $\observations_0 = \observations$
	
	\For{t=1, ..., T}{
		Compute $\epsilon$ and sample $u \sim \mathcal{U}(0,1)$
		
		\If{$\epsilon > u$}{(Observe) \\
			1. Observe new observations $(\mat{x}_t, c_t, \mat{y}_t)$.\\
			2. Augment $\observations = \observations \cup \{(\mat{x}_t, c_t, \mat{y}_t, )\}$.\\
			3. Update prior of the causal \gptext (\eq \eqref{eq:gp_model}). 
		}
		\Else{(Intervene)\\
			1. Compute $EI^s(\mat{x})/Co(\mat{x})$ for each element $s \in \expset$ (\eq \eqref{eq:causal_acquisition}).  \\
			2. Obtain the optimal interventional set-value pair $(s^\star, \alpha^\star)$.\\
			3. Intervene on the system.\\
			4. Update posterior of the causal \gptext.
		}
	}
	Return the optimal value $\hatexpectation{}{\mat{Y}^\star|\DO{\mat{X}^\star_s}{\x^\star_s}}$ in $\true_T$ and the corresponding $\mat{X}_s^\star, \x_s^\star$.
	\caption{Causal Bayesian Optimization - \cbo}
	\label{alg:ceo}
	\end{algorithm}

\section{Experiments}
We test our algorithm on a synthetic setting and on two real-world applications for which a \DAG is available and can be used as a simulator. We run \cbo to explore both $\missets$ and $\pomissets$ and show how the optimal intervention set, intervention values and cost incurred to achieve the optimum change depending on the \DAG and the \sem. 
For all variables in $\mat{X}$ and their combinations, we assume to have data from previous interventions which we denote by $\true = \{(\x_s^i, \expectation{}{Y|\DO{\mat{X}_s^i}{\x_s^i}})\}_{i=1, s=1}^{P, |\expset|}$. Typically $P$ is very small and interventions are prohibitive to implement.

\textbf{Baselines:}
We compare \cbo against a standard \bo algorithm, in which all variables are intervened upon, and a \cbo version where a standard \gptext prior given by $p(f(\x_s)) = \gp (0, k_{\rbf}(\x_s, \x'_s))$ is used. 

\textbf{Performance measures:}
We run \cbo with different initializations of $\true$ and report the average convergence performances together with standard errors. In the synthetic setting, we consider three different cost configurations: equal unit cost per node, different fix costs per node and variable costs per node. The total cost at each optimization step is computed as the sum of the cost for each intervened node. We show the results for equal unit cost per node and report the full comparison in the supplement. 

\subsection{Toy Experiment}
\begin{figure}[t]
	\centering
	\includegraphics[width=0.45\textwidth]{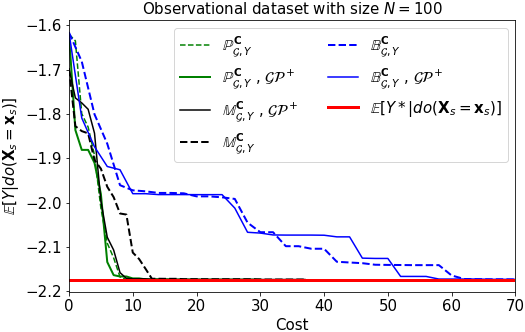}
	\caption{Toy example. Convergence of \cbo and standard \bo across different initializations of $\dataset^I$. The red line gives the optimal $Y^*$ when intervening on sets in $\missets$, $\pomissets$ or $\boset$. Solid lines give \cbo results when using the causal \gptext model which is denoted by $\mathcal{GP}^+$. Dotted line correspond to \cbo with a standard \gptext prior model $p(f(\x_s)) = \gp (0, k_{\rbf}(\x_s, \x'_s))$. See the supplement for standard deviations (\fig (1)).}
	\label{fig:toy_results}
	\vspace{-0.3cm}
\end{figure}

We show the convergence results for \cbo and competing algorithms for the toy example described in the text. For this experiment we set $N=100$ and $P=3$. Given the \sem in \fig \ref{fig:toy_1}, the optimal configuration is $(X^\star_s, x^\star_s)= (Z, -3.20)$. \cbo converges to the optimum faster than \bo which requires to intervene on all nodes and it is thus twice more expensive (\fig \ref{fig:toy_results}). 

\begin{figure*}[t]
	\centering
	\includegraphics[width=8cm,height=5.15cm]{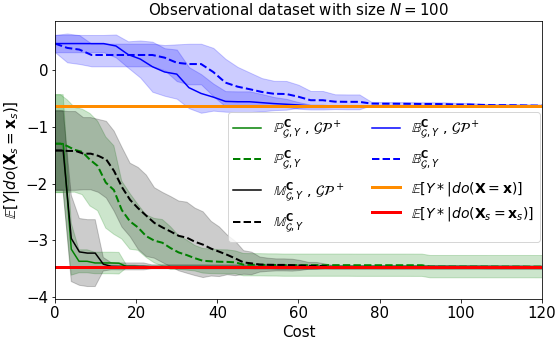}
	\includegraphics[width=8cm,height=5.15cm]{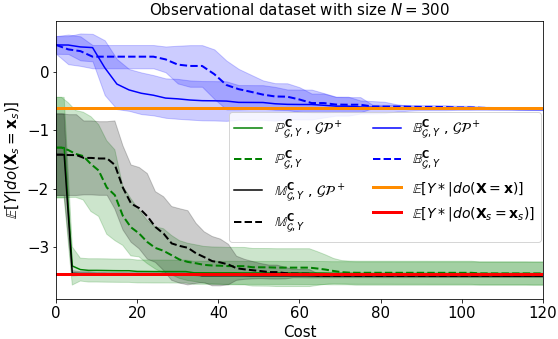} 
	\caption{Synthetic example. Convergence of \cbo and \bo across different initialization of $\dataset^I$. The orange line gives the optimal $Y^*$ when intervening on $\boset$. The red line gives the optimal $Y^*$ when intervening on sets in $\missets$ or $\pomissets$. Solid lines give \cbo results when using the causal \gptext model, denoted by $\mathcal{GP}^+$, while dotted lines correspond to \cbo with a standard \gptext prior model. Shaded areas are $\pm$ standard deviation.}
	\label{fig:synthetic_resesults}
	\vspace{-0.3cm}
\end{figure*}

\subsection{Synthetic Experiment}
We test the algorithm on the \DAG given in \fig \ref{fig:dag_related_methods} (top). This \DAG includes unobserved confounders, non-manipulative variables and requires to apply both front-door and back-door adjustment formulas to estimate  the causal effects. We set $P=10$ and test different values of $N$. The \expset{s} are $\missets = \{\emptyset, \{B\}, \{D\}, \{E\}, \{B,D\}, \{B,E\}, \{D, E\}\}$, $\pomissets = \{\emptyset, \{B\}, \{D\}, \{E\}, \{B,D\}, \{D, E\}\}$ and $\boset = \{B,D,E\}$. All the intervention sets in $\missets$ and $\pomissets$ include a maximum of two variables. On the contrary, \bo only considers interventions on three variables thus increasing the dimensionality of the problem by one. The \sem,  the \textit{do}-calculus computations and further details about the experimental settings are given in the supplement.  

\fig \ref{fig:synthetic_resesults} shows how \cbo outperforms standard \bo and achieves the best performance when the causal \gptext model is used.  There are two main reasons why \bo leads to a suboptimal solution.  
Firstly, $\expectation{}{Y| \DO{B}{b},\DO{D}{d}, \DO{E}{e}} = \expectation{}{Y| \DO{D}{d},\DO{E}{e}}$. This means that the same outcome can be achieved by intervening on $\{D, E\}$ at a significantly lower cost. 

Secondly, although it may seem counterintuitive, intervening only on a subset of variables leads to better outcomes. Manipulating all variables breaks the causal links between them and blocks the propagation of causal effects in the graph. In this example, intervening on $B,E,D$ blocks the causal effect of $B$ on $Y$. Manipulating only $B$ leads to a propagation of its causal effect through $D$ and $E$. Given the \sem, $\expectation{}{Y|\text{do}(B,=b, D=d,E=e)} < \expectation{}{Y|\text{do}(B,=b, D=d)}$ $\forall \quad b \in D(B), d \in D(D), e \in D(E)$.
Indeed, setting the level of $B$ makes $D$ and $E$ take values outside of their interventional domains $D(D)$ and $D(E)$ thus leading to function values not achievable in \bo. 
Furthermore, the causal \gptext prior determines the locations of the function evaluations thus reducing the number of steps required to find the optimum. As expected, the benefit of incorporating $\observations$ into the prior becomes more evident when $N$ increases. The optimal configuration for this setting is $(\mat{X}^\star_s, \x^\star_s)=(\{B, D\}, (-5.0,3.28))$.

\subsection{Example in Ecology}
\begin{figure}[t]
	\centering
	\includegraphics[width=0.48\textwidth]{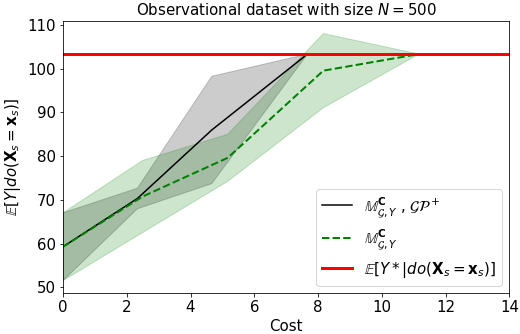}
	\caption{\nec example. Convergence of \cbo across different initialization of $\dataset^I$, with and without causal \gptext prior. The red line gives the optimal $Y^\star$ when intervening on $\missets$.}
	\label{fig:nec_results}
	\vspace{-0.3cm}
\end{figure}

We apply \cbo to a large-scale optimization problem in ecology. We consider the issue of maximizing the net coral ecosystem calcification (\nec) in the Bermuda given a set of environmental variables. The causal graph (\fig 4 in the supplement) is taken from \cite{courtney2017environmental} and modified so as to avoid directed cycles. We consider a subset of 5 variables as manipulative, that is $\mat{X}=\{\text{Nut}, \Omega_{A}, Chl\alpha, \text{\ta}, \text{\dic}\}$, and assume to be able to intervene \textit{contemporaneously} on a maximum of 3 variables. Given these assumptions and the \DAG, $\missets$ includes the single variable interventions and all the 2 and 3 variables interventions that can be performed selecting variables in $\mat{X}$. The cardinality of $\missets$ is thus 25. Notice that the size of $\boset = \{\text{Nut}, \Omega_{A}, Chl\alpha, \text{\ta}, \text{\dic}\}$ is greater than 3 thus \bo is not a viable strategy for this application. We first construct a simulator by fitting a linear \sem with the 50 observations provided by \cite{nec_data}. 
We then use the simulator to generate $N=500$ observations and $P=1$ initial interventional data points. We set the interventional domains to $D(\text{Nut}) = [-2,5]$, $D(\Omega_{A}) = [2,4]$, $D(Chl\alpha) = [0.3,0.4]$, $D(\text{\ta}) = [2200,2550]$ and $D(\text{\dic}) = [1950,2150]$. We run \cbo on $\missets$, with and without the causal \gptext prior. We found \cbo to successfully explore $\missets$, especially when the causal \gptext prior is used (\fig \ref{fig:nec_results}). The optimal intervention is $(\mat{X}^\star_s, \x^\star_s)=(\{\Omega_{A}, \text{\ta},\text{\dic}\}, (2,2550,1950))$.

\subsection{Example in Healthcare}
Finally, we apply our method to an example in healthcare.
The \DAG (\fig 3 in the supplement) is taken from \cite{Thompson} and \cite{ferro2015use} which is used to model the causal effect of statin drugs on the levels of prostate specific antigen (\psa). Our goal is to minimize \psa by intervening on statin and aspirin usage. $\dataset^O$ consists of $N=500$ instances sampled from the simulator while $P=3$. Given the causal structure, $\missets = \{\emptyset,\{\text{aspirin}\}, \{\text{statin}\}, \{\text{aspirin}, \text{statin}\}\}$ while $\pomissets = \{\{\text{aspirin}, \text{statin}\}\}$. We set the domain $D(\text{aspirin})=D(\text{statin})=[0.0,1.0]$ and run \cbo on both $\missets$ and $\pomissets$. We found the optimal intervention to be $(\mat{X}^\star_s, \x^\star_s)=(\{\text{aspirin}, \text{statin}\}, (0.0,1.0))$. 
This experiment shows how \cbo can help doctors, and decision-makers in general, to find optimal interventions in real-life scenarios based on simulators and thus avoiding expensive and invasive interventions.

\section{Conclusions}

This paper formalizes the problem of globally optimizing a variable that is part of a causal model in which a sequence of interventions can be performed. We propose a Causal Bayesian Optimization (\cbo) algorithm which solves the global optimization problem exploring a set of potentially optimal sets. This is achieved via a causal expected improvement acquisition function and an $\epsilon$-greedy policy solving the emerging observation-intervention trade off. In addition, we formulate a causal \gptext model which allows to integrate observational and interventional data via \dotext-calculus. 
We show the benefits of the proposed approach in a variety of settings characterized by different causal graph structures. Our results demonstrate how \cbo outperforms \bo and reaches the global optimum after a significantly lower number of optimization steps. 
Future work will focus on reducing the number of causal \gptext{s} used by \cbo to explore $\expset$. For instance, this could be achieved through a multi-output \gptext model where different outputs correspond to different interventional sets. 
In addition, we will work on extending \cbo to multi-objective optimization. Finally, we will focus on combining the proposed framework with a causal discovery algorithm so as to account for uncertainty in the graph structure.

\bibliographystyle{apalike}
\bibliography{references}

\begin{thebibliography}{}

\bibitem[Andersson, 2018]{nec_data}
Andersson, A., B.~N. (2018).
\newblock In situ measurements used for coral and reef-scale calcification
  structural equation modeling including environmental and chemical
  measurements, and coral calcification rates in bermuda from 2010 to 2012
  (beacon project).
\newblock {\em Biological and Chemical Oceanography Data Management Office
  (BCO-DMO). Dataset version 2018-03-02.
  http://lod.bco-dmo.org/id/dataset/720788}.

\bibitem[Bareinboim et~al., 2015]{bareinboim2015bandits}
Bareinboim, E., Forney, A., and Pearl, J. (2015).
\newblock Bandits with unobserved confounders: A causal approach.
\newblock In {\em Advances in Neural Information Processing Systems}, pages
  1342--1350.

\bibitem[Buesing et~al., 2018]{buesing2018woulda}
Buesing, L., Weber, T., Zwols, Y., Racaniere, S., Guez, A., Lespiau, J.-B., and
  Heess, N. (2018).
\newblock Woulda, coulda, shoulda: Counterfactually-guided policy search.
\newblock {\em arXiv preprint arXiv:1811.06272}.

\bibitem[Cochran and Cox, 1957]{cochran1957experimental}
Cochran, W. and Cox, G. (1957).
\newblock Experimental design. john willey and sons.
\newblock {\em Inc., New York, NY}.

\bibitem[Courtney et~al., 2017]{courtney2017environmental}
Courtney, T.~A., Lebrato, M., Bates, N.~R., Collins, A., De~Putron, S.~J.,
  Garley, R., Johnson, R., Molinero, J.-C., Noyes, T.~J., Sabine, C.~L., et~al.
  (2017).
\newblock Environmental controls on modern scleractinian coral and reef-scale
  calcification.
\newblock {\em Science advances}, 3(11):e1701356.

\bibitem[Ferro et~al., 2015]{ferro2015use}
Ferro, A., Pina, F., Severo, M., Dias, P., Botelho, F., and Lunet, N. (2015).
\newblock Use of statins and serum levels of prostate specific antigen.
\newblock {\em Acta Urol{\'o}gica Portuguesa}, 32(2):71--77.

\bibitem[Foerster et~al., 2018]{foerster2018counterfactual}
Foerster, J.~N., Farquhar, G., Afouras, T., Nardelli, N., and Whiteson, S.
  (2018).
\newblock Counterfactual multi-agent policy gradients.
\newblock In {\em Thirty-Second AAAI Conference on Artificial Intelligence}.

\bibitem[Glymour et~al., 2019]{glymour2019review}
Glymour, C., Zhang, K., and Spirtes, P. (2019).
\newblock Review of causal discovery methods based on graphical models.
\newblock {\em Frontiers in Genetics}, 10.

\bibitem[Guo et~al., 2018]{guo2018survey}
Guo, R., Cheng, L., Li, J., Hahn, P.~R., and Liu, H. (2018).
\newblock A survey of learning causality with data: Problems and methods.
\newblock {\em arXiv preprint arXiv:1809.09337}.

\bibitem[Jones et~al., 1998]{jones1998efficient}
Jones, D.~R., Schonlau, M., and Welch, W.~J. (1998).
\newblock Efficient global optimization of expensive black-box functions.
\newblock {\em Journal of Global optimization}, 13(4):455--492.

\bibitem[Lattimore et~al., 2016]{lattimore2016causal}
Lattimore, F., Lattimore, T., and Reid, M.~D. (2016).
\newblock Causal bandits: Learning good interventions via causal inference.
\newblock In {\em Advances in Neural Information Processing Systems}, pages
  1181--1189.

\bibitem[Lee and Bareinboim, 2018]{lee2018structural}
Lee, S. and Bareinboim, E. (2018).
\newblock Structural causal bandits: where to intervene?
\newblock In {\em Advances in Neural Information Processing Systems}, pages
  2568--2578.

\bibitem[Lee and Bareinboim, 2019]{lee2019structural}
Lee, S. and Bareinboim, E. (2019).
\newblock Structural causal bandits with non-manipulable variables.
\newblock Technical report, Technical Report R-40, Purdue AI Lab, Department of
  Computer Science, Purdue.

\bibitem[Lu et~al., 2018]{lu2018deconfounding}
Lu, C., Sch{\"o}lkopf, B., and Hern{\'a}ndez-Lobato, J.~M. (2018).
\newblock Deconfounding reinforcement learning in observational settings.
\newblock {\em arXiv preprint arXiv:1812.10576}.

\bibitem[Ortega and Braun, 2014]{ortega2014generalized}
Ortega, P.~A. and Braun, D.~A. (2014).
\newblock Generalized thompson sampling for sequential decision-making and
  causal inference.
\newblock {\em Complex Adaptive Systems Modeling}, 2(1):2.

\bibitem[Pearl, 1995]{pearl1995causal}
Pearl, J. (1995).
\newblock Causal diagrams for empirical research.
\newblock {\em Biometrika}, 82(4):669--688.

\bibitem[Pearl, 2000]{pearl2000causality}
Pearl, J. (2000).
\newblock {\em Causality: models, reasoning and inference}, volume~29.
\newblock Springer.

\bibitem[Rasmussen, 2003]{rasmussen2003gaussian}
Rasmussen, C.~E. (2003).
\newblock Gaussian processes in machine learning.
\newblock In {\em Summer School on Machine Learning}, pages 63--71. Springer.

\bibitem[Shahriari et~al., 2015]{shahriari2015taking}
Shahriari, B., Swersky, K., Wang, Z., Adams, R.~P., and De~Freitas, N. (2015).
\newblock Taking the human out of the loop: A review of bayesian optimization.
\newblock {\em Proceedings of the IEEE}, 104(1):148--175.

\bibitem[Srinivas et~al., 2012]{srinivas2012information}
Srinivas, N., Krause, A., Kakade, S.~M., and Seeger, M.~W. (2012).
\newblock Information-theoretic regret bounds for gaussian process optimization
  in the bandit setting.
\newblock {\em IEEE Transactions on Information Theory}, 58(5):3250--3265.

\bibitem[Thompson, 2019]{Thompson}
Thompson, C. (2019).
\newblock Causal graph analysis with the causalgraph procedure.
\newblock
  \url{https://www.sas.com/content/dam/SAS/support/en/sas-global-forum-proceedings/2019/2998-2019.pdf}.

\bibitem[Tokic, 2010]{tokic2010adaptive}
Tokic, M. (2010).
\newblock Adaptive $\varepsilon$-greedy exploration in reinforcement learning
  based on value differences.
\newblock In {\em Annual Conference on Artificial Intelligence}, pages
  203--210. Springer.

\bibitem[Wang et~al., 2016]{wang2016bayesian}
Wang, Z., Hutter, F., Zoghi, M., Matheson, D., and de~Feitas, N. (2016).
\newblock Bayesian optimization in a billion dimensions via random embeddings.
\newblock {\em Journal of Artificial Intelligence Research}, 55:361--387.

\end{thebibliography}

\end{document}


\twocolumn[

\aistatstitle{Supplementary Material for ``Causal Bayesian Optimization''}

\aistatsauthor{Virginia Aglietti \And  Xiaoyu Lu \And Andrei Paleyes \And Javier Gonz\' alez}

\aistatsaddress{
	University of Warwick\\
	The Alan Turing Institute\\ 
	\url{V.Aglietti@warwick.ac.uk}\\ \And  
	Amazon \\
	Cambridge, UK \\
	\url{luxiaoyu@amazon.com}\\ \And 
	Amazon \\
	Cambridge, UK \\ 
	\url{paleyes@amazon.com}\\ \And 
	Amazon \\
	Cambridge, UK\\
	\url{gojav@amazon.com} }

\section{Derivations of \textit{do}-calculus for the synthetic experiment}

\subsection{$Do(B=b)$  }

\begin{align*}
p(y|do(B=b)) &= \int p(y|c, do(B=b))p(c|B=b)dc \\
&= \int p(y|do(C=c), do(B=b))p(C=c|B=b)dc  \;\;\; (Y \indep C | B \textit{ in } \mathcal{G}_{\bar{B} \underline{C}}) \\
&=\int p(y|do(C=c))p(c|B=b)dc \;\;\; (Y \indep B|C \textit{ in } \mathcal{G}_{\bar{B},\bar{C}})   \\
&= \int p(y|b', do(C=c))p(b'| do(C=c))p(c|B=b)db'dc \\
&= \int p(y|b', C=c) p(b') p(c|B=b) db'dc  \;\;\; (Y \indep C|B   \textit{ in }  \mathcal{G}_{\bar{B},\underline{C}})
\end{align*}

\subsection{$Do(D=d)$  }
\begin{align*}
p(y|do(D=d)) &= \int p(y|c, do(D=d)) p(c|do(D=d))db    \\
&= \int p(y|c, D=d) p(c) dc \;\;\; (Y \indep D |C \textit{ in }   \mathcal{G}_{\underline{D}})
\end{align*}

\subsection{$Do(E=e)$  }
\begin{align*}
p(y|do(E=e)) &= \int p(y|a,c,do(E=e)) p(a,c |do(E=e)) dadc \\
&= \int p(y|a,c,E=e) p(a)p(c) dadc  \;\;\; (Y \indep E |A,C  \textit{ in }   \mathcal{G}_{\underline{E}})
\end{align*}

\subsection{$Do(B=b, D=d)$  }
\begin{align*}
p(y|do(B=b), do(D=d)) &= \int p(y|do(B=b), c, do(D=d)) p(c|do(B=b),do(D=d)) dc \\
&= \int p(y|do(B=b), do(C=c), do(D=d)) p(c|B=b) dc  \;\;\;  (Y \indep C |B,D \textit{ in }  \mathcal{G}_{\underline{C} \bar{B} \bar{D}})  \\
&= \int p(y|do(C=c), do(D=d)) p(c|B=b) dc \;\;\; (Y \indep B|C,D  \textit{ in }  \mathcal{G}_{\bar{B} \bar{C} \bar{D}})  \\
&= \int p(y|b', do(C=c), do(D=d)) p(b'|do(C=c), do(D=d)) p(c|B=b)dcdb' \\
&= \int p(y|b', C=c, do(D=d)) p(b')p(c|B=b) dcdb' \;\;\; (Y \indep C|B,D \textit{ in }    \mathcal{G}_{\bar{B} \bar{D} \underline{C}})  \\
&= \int p(y|b',C=c, D=d) p(b') p(c|B=b) dcdb'  \;\;\; (Y \indep D|B,C \textit{ in }    \mathcal{G}_{\underline{D}}) 
\end{align*}

]
\onecolumn

\subsection{$Do(B=b, E=e)$  }
\begin{align*}
p(y|do(B=b), do(E=e)) &= \int p(y|do(B=b), c, do(E=e)) p(c|B=b) dc \\
&= \int p(y|do(B=b), do(C=c), do(E=e)) p(c|B=b)dc   \;\;\; (Y \indep C|B,E \textit{ in }    \mathcal{G}_{\bar{B} \bar{E} \underline{C}}) \\
&= \int p(y|do(C=c), do(E=e)) p(c|B=b)dc  \;\;\; (Y \indep B|C,E \textit{ in }    \mathcal{G}_{\bar{C} \bar{E} \bar{B}}) \\
&= \int p(y|do(C=c), do(E=e), b')p(b'|do(C=c), do(E=e)) p(c|B=b) db'dc \\
&= \int p(y|C=c, do(E=e), b')p(b') p(c|B=b) db'dc   \;\;\; (Y \indep C|B,E \textit{ in }    \mathcal{G}_{\bar{E} \underline{C}}) \\
&= \int p(y|a, C=c, do(E=e), b')p(a|C=c, do(E=e), b')p(b') p(c|B=b) db'dc da \\
&=  \int p(y|a,b',C=c, E=e)p(a)p(b')p(c|B=b) db'dcda   \;\;\; (Y \indep E|A, B,C \textit{ in }    \mathcal{G}_{ \underline{E}}) \\
\end{align*}

\subsection{$Do(D=d, E=e)$  }
\begin{align*}
p(y|do(D=d),do(E=e)) &= \int p(y|a,c,do(D=d), do(E=e)) p(a,c |do(D=d),do(E=e))dadc \\
&= \int p(y|a,c,D=d, E=e) p(a)p(c) dadc  \;\;\; (Y \indep (D,E)|A, C \textit{ in }    \mathcal{G}_{ \underline{D, E}}) \\
\end{align*}

\subsection{$Do(B=b, D=d, E=e)$  }
\begin{align*}
p(y|do(B=b), do(D=d),do(E=e)) = p(y|do(D=d),do(E=e))  \;\;\; (Y \indep B|D, E \textit{ in }    \mathcal{G}_{ \bar{D}, \bar{E}, \bar{B}})
\end{align*}

\section{\sem for the synthetic experiment}

The \sem for the synthetic example is:

\begin{align*}
U_1 &= \epsilon_{YA} \\
U_2 &= \epsilon_{YB} \\
F &= \epsilon_{F} \\
A &= F^2  + U_1 + \epsilon_A\\
B &=  U_2 + \epsilon_B\\
C &= \exp(-B) + \epsilon_C \\
D &= \exp(-C)/10. + + \epsilon_D \\
E &= \text{cos}(A) + C/10 + \epsilon_E \\
Y &= \text{cos}(D) + \text{sin}(E) + U_1 + U_2  \epsilon_y \\
\end{align*}

\section{Cost configurations}
\label{ref:costs}
Denote by $Co(\mat{X}, \mat{x})$ the cost of intervening on node $\mat{X}$ at the value $\mat{x}$. For the toy example and the real-data examples we consider fix unit cost across nodes. For the synthetic example we consider three possible cost configurations: equal fix costs across nodes, different fix costs across nodes and variable costs across nodes. These are set to:
\begin{enumerate}
	\item Fix equal costs: $Co(B, b) = Co(D, d) = Co(E, e) = Co(F, f) = 1$.
	\item Fix different costs: $Co(B, b) = 10$, $Co(D, d)=5$, $Co(E, e)=20$ and $Co(F, f)=3$.
	\item Variable costs: $Co(B, b) = 10+|b|$, $Co(D,d)=5+|d|$, $Co(E,e)=20+|e|$ and $Co(F,f)=3+|f|$.
\end{enumerate}

\section{Additional synthetic results}

\begin{figure}[t]
	\centering
	\includegraphics[width=0.45\textwidth]{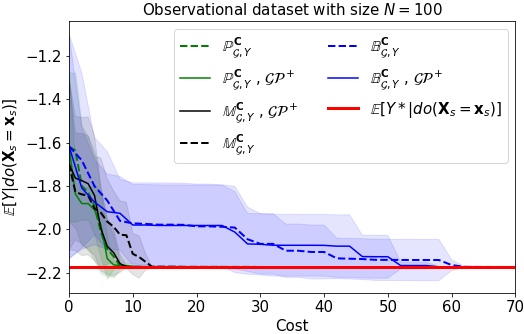}
	\caption{Toy example. Convergence of \cbo and standard \bo across different initializations of $\dataset^I$. The red line gives the optimal $Y^*$ when intervening on sets in $\missets$, $\pomissets$ or $\boset$. Solid lines give \cbo results when using the causal \gptext model which is denoted by $\mathcal{GP}^+$. Dotted line correspond to \cbo with a standard \gptext prior model $p(f(\x_s)) = \gp (0, k_{\rbf}(\x_s, \x'_s))$. Shaded areas are $\pm$ standard deviation.}
	\label{fig:toy_results2}
\end{figure}

In \fig \ref{fig:toy_results2} we show the results for the toy experiment across different initialization of $\dataset^I$.

\begin{figure*}[t]
	\centering
	\begin{center}
		\includegraphics[width=6cm,height=4.15cm]{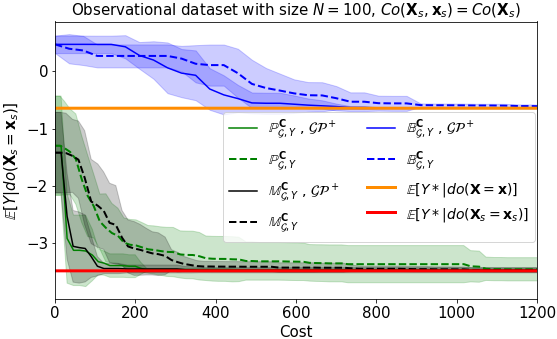}
		\includegraphics[width=6cm,height=4.15cm]{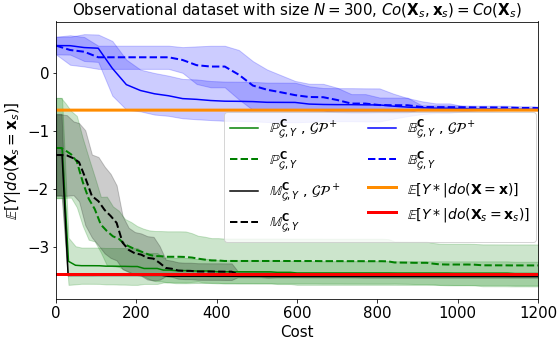} 	
		\includegraphics[width=6cm,height=4.15cm]{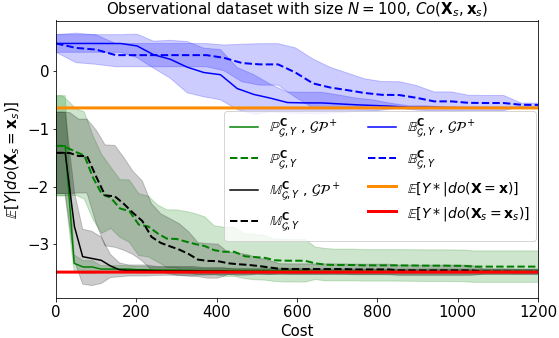} 	
		\includegraphics[width=6cm,height=4.15cm]{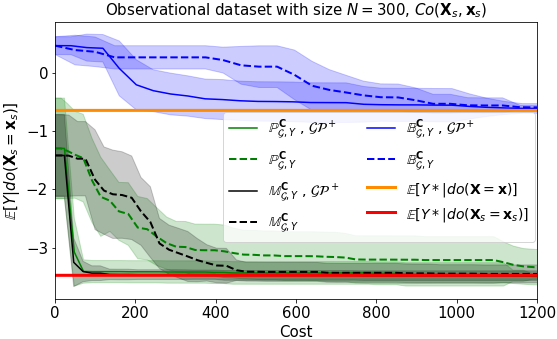} 	
	\end{center}
	
	\caption{Synthetic example. Convergence of \cbo and standard \bo. The orange line gives the optimal $Y^*$ when intervening on $\boset$. The red line gives the optimal $Y^*$ when intervening on sets in $\missets$ or $\pomissets$. Solid lines give \cbo results when using the causal \gptext model which is denoted by $\mathcal{GP}^+$. Dotted line correspond to \cbo with a standard \gptext prior model. \textit{Upper left}: option (2) in \S\ref{ref:costs}, $N=100$. \textit{lower left}: option (3) in \S\ref{ref:costs}, $N=100$.  \textit{Upper right}: option (2) in \S\ref{ref:costs}, $N=300$.  \textit{Lower right}: option (3) in \S\ref{ref:costs}, $N=300$.}
	\label{fig:synthetic_resesults}
\end{figure*}

In \fig \ref{fig:synthetic_resesults} we show the results for the synthetic experiment across different cost structures and values of $N$.

\section{Example in Healthcare}
\begin{figure}
	\centering
	\centering
	\begin{tikzpicture}[shorten >=1pt, node distance=2mm, on grid]
	\node[obs, pattern=dots] (X_1) {$\textbf{\bmi}$} ; %
	\node[obs, below = 3cm of X_1, pattern=dots] (X_2) {$\textbf{Cancer}$} ; %
	\node[obs, above left = 3cm of X_2,  pattern=dots] (X_3) {$\textbf{Age}$} ; %
	\node[obs, above right = 3cm of X_2] (X_4) {$\textbf{Aspirin}$} ; %
	\node[obs, below = 3cm of X_3] (X_5) {$\textbf{Statin}$} ; %
	\node[obs, right = 4cm of X_5, ultra thick] (Y) {$\textbf{\psa}$} ; %
	\edge {X_1} {X_2} ;
	\edge {X_1} {X_4} ;
	\edge {X_1} {X_5} ;
	\edge {X_1} {Y} ;
	\edge {X_3} {X_1} ;
	\edge {X_3} {X_2} ;
	\edge {X_3} {X_4} ;
	\edge {X_3} {X_5} ;
	\edge {X_4} {X_2} ;
	\edge {X_4} {Y} ;
	\edge {X_5} {X_2} ;
	\edge {X_5} {Y} ;
	\edge {X_2} {Y} ;
	\path[->] (X_3) edge[bend left=22] (Y);
	\end{tikzpicture}
	\caption{Causal graph of \psa level. Shaded nodes represent variables which can be intervened and dotted nodes represent non-manipulative variables. The target variable \psa is denoted with a thick shaded node. }
	\label{fig:dag_psa}
\end{figure}
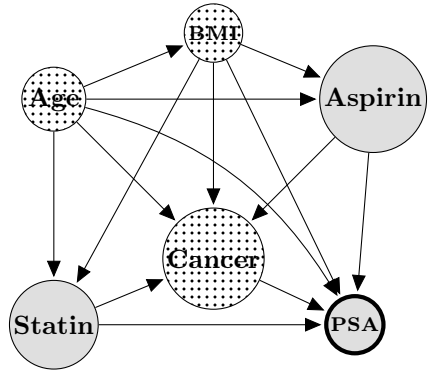

The \DAG describing the causal relationships between statin drugs and \psa \citep{Thompson, ferro2015use} is given in \fig \ref{fig:dag_psa}. The \sem for this example is:

\begin{align*}
age &= \mathcal{U}(55,75) \\
bmi &= \mathcal{N}(27.0 - 0.01 \times age, 0.7) \\
aspirin &= \sigma(-8.0 + 0.10\times age + 0.03 \times bmi) \\
statin &= \sigma(-13.0 + 0.10\times age + 0.20\times bmi)\\
cancer &=  \sigma(2.2 - 0.05\times age + 0.01 \times bmi - 0.04 \times statin + 0.02\times aspirin) \\
Y &=  \mathcal{N}(6.8 + 0.04 \times age - 0.15 \times bmi - 0.60 \times statin + 0.55 \times aspirin + 1.00 \times cancer, 0.4) \\
\end{align*}

where $\mathcal{U}(a,b)$ denotes a uniform random variable with parameters $a$ and $b$, $\mathcal{N}(m,s)$ represents a normal random variable with mean $m$ and standard deviation $s$ and $\sigma$ denotes the sigmoidal function computed as $\sigma(x) = \frac{1}{1+e^{-x}}$.

\section{Example in Ecology}
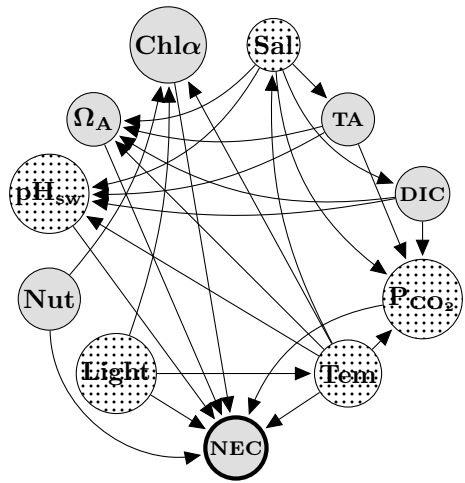
\begin{figure}
	\centering
	\centering
	\begin{tikzpicture}[shorten >=1pt, node distance=3mm, on grid]
	\node[obs, ultra thick] (Y) {$\textbf{\nec}$} ; %
	\node[obs, above left = 1.4cm of Y, xshift=-0.6cm, pattern=dots] (L) {$\textbf{Light}$} ; %
	\node[obs, above left = 1.4cm of L,xshift=0.1cm,] (N) {$\textbf{Nut}$} ; %
	\node[obs, above = 1.4cm of N,  pattern=dots] (P) {$\mat{pH_{sw}}$} ; %
	\node[obs, above right = 1.4cm of P, xshift =-0.4cm] (O) {$\mat{\mat\Omega_{A}}$} ; %
	\node[obs, above right = 1.4cm of O] (C) {$\mat{Chl}\vect{\alpha}$} ; %
	\node[obs, right = 1.4cm of C,  pattern=dots] (S) {$\textbf{Sal}$} ; %
	\node[obs, below right = 1.4cm of S] (T) {$\textbf{\ta}$} ; %
	\node[obs, below right = 1.4cm of T] (D) {$\textbf{\dic}$} ; %
	\node[obs, below = 1.4cm of D,  pattern=dots] (CO) {$\mat{P_{CO_2}}$} ; %
		\node[obs, below left = 1.4cm of CO,  pattern=dots] (TE) {$\textbf{Tem}$} ; %
		
		\edge {L} {Y} ;
		\edge {L} {TE} ;
		\edge {TE} {Y} ;
		\edge {C} {Y} ;
		\edge {O} {Y} ;
		\edge {P} {Y} ;
\path[->] (S) edge[bend right=0] (T);
\path[->] (S) edge[bend right=25] (D);
\path[->] (S) edge[bend right=25] (CO);
\path[->] (S) edge[bend left=25] (O);
\path[->] (S) edge[bend left=25] (P);
\path[->] (T) edge[bend left=15] (O);
\path[->] (T) edge[bend left=15] (P);
\path[->] (T) edge[bend right=0] (CO);
\path[->] (D) edge[bend left=20] (O);
\path[->] (D) edge[bend left=10] (P);
\path[->] (D) edge[bend left=0] (CO);
\path[->] (TE) edge[bend left=0] (CO);
\path[->] (TE) edge[bend left=15] (S);
\path[->] (TE) edge[bend left=0] (C);
\path[->] (TE) edge[bend left=0] (O);
\path[->] (TE) edge[bend left=0] (P);
\path[->] (N) edge[bend right=15] (C);
\path[->] (N) edge[bend right=50] (Y);
\path[->] (CO) edge[bend right=30] (Y);
\path[->] (L) edge[bend right=10] (C);
	\end{tikzpicture}
	\caption{\DAG of \nec level. Shaded nodes represent manipulative variables. Dotted nodes represent non-manipulative variables. The target variable \nec is denoted with a thick shaded node. A description of the variables can be found in the supplement.}
	\label{fig:dag_nec}
	\vspace{-0.45cm}
\end{figure}

The \DAG describing the causal relationships between a set of environmental variables and \nec \citep{courtney2017environmental} is given in \fig \ref{fig:dag_nec}. The variables included in the \DAG are:

\begin{itemize}
	\item $Chl\alpha$: sea surface chlorophyll a;
	\item Sal: sea surface salinity;
	\item \ta: seawater total alkalinity;
	\item \dic: seawater dissolved inorganic carbon;
	\item $P_{CO_2}$: seawater $P_{CO_2}$;
	\item Tem: bottom temperature;
	\item \nec: net ecosystem calcification;
	\item Light: bottom light levels;
	\item Nut: PC1 of NH4, NiO2+NiO3, SiO4;
	\item $pH_{SW}$: seawater pH;
	\item $\Omega_A$: seawater saturation with respect to aragonite.
\end{itemize}

See \cite{nec_data} for more details.
	
\bibliographystyle{apalike}
\bibliography{references}